\documentclass[runningheads]{llncs}
\usepackage{graphicx}
\usepackage[utf8]{inputenc} 
\usepackage[T1]{fontenc}    
\usepackage{hyperref}       
\usepackage{booktabs}       
\usepackage{amsfonts}       
\usepackage{nicefrac}       
\usepackage{microtype}      
\usepackage{amsmath}
\usepackage{cleveref}
\usepackage{subcaption}
\usepackage{diagbox}

\begin{document}
\title{Constraint-Based Visual Generation\thanks{This is a post-peer-review, pre-copyedit version of an article published in LNCS, volume 11729. The final authenticated version is available online at: \texttt{https://doi.org/10.1007/978-3-030-30508-6\_45}}}

%
\author{Giuseppe Marra\inst{1,2} \and
Francesco Giannini\inst{2} \and \\
Michelangelo Diligenti\inst{2} \and Marco Gori\inst{2}}
\authorrunning{G. Marra et al.}
%
\institute{Department of Information Engineering, University of Florence, ITALY
\email{g.marra@unifi.it}\\	
	 \and
Department of Information Engineering and Mathematical Sciences, \\ University of Siena, ITALY \\ \email{\{fgiannini,diligmic,marco\}@diism.unisi.it}}
\maketitle              
\begin{abstract}
In the last few years the systematic adoption of deep learning to visual generation has produced impressive results that, amongst others, definitely benefit from the massive exploration of convolutional architectures. In this paper, we propose a general approach to visual generation that combines learning capabilities with logic descriptions of the target to be generated. The process of  generation is regarded as a constrained satisfaction problem, where the constraints describe a set of properties that characterize the target. Interestingly, the constraints can also involve logic variables, while all of them are converted into real-valued functions by means of the t-norm theory. We use deep architectures to model the involved variables, and propose a computational scheme where the learning process carries out a satisfaction of the constraints. We propose some examples in which the theory can  naturally be used, including the modeling of GAN and auto-encoders, and report promising results in image  translation of human faces.

\keywords{GANs  \and Declarative Language \and Visual Generation \and FOL.}
\end{abstract}

\section{Introduction}

\label{sec:intro}
Generative Adversarial Networks (GANs)~\cite{goodfellow2014generative} have achieved impressive results in image generation. By taking inspiration from the Turing test, a generator function is asked to fool a discriminator function which, in turn, tries to distinguish real samples from generated ones. GANs are known to generate very realistic images when trained properly.
A special generation task is image-to-image translation, which learns to map each image for an input domain into an image in a (possibly different) output domain.
In most real-world domains, there are no pairs of examples showing how to translate an image into a corresponding one in another domain, yielding the so called UNsupervised Image-to-image Translation (UNIT) problem.
In an UNIT problem, two independent sets of images belonging to two different domains (e.g. cats-dogs, male-female, summer-winter, etc.) are given and the task is to translate an image from one domain into the corresponding image in the other domain, even though there exist no paired examples showing this mapping.
Unfortunately, estimating a joint distribution of the images in the two domains from the distributions in the original single domains is known to have infinite possible solutions.
Therefore, one possible strategy consists in mapping pairs of corresponding images to the same latent space using auto-encoders and then learning to reconstruct an image from its representation in latent space. Combining auto-encoders with GANs has been proposed in~\cite{rosca2017variational,li2017alice} and outstanding results on image translation have been reported by~\cite{zhu2017unpaired,liu2016coupled,liu2017unsupervised}.

This paper proposes a general approach to visual generation and translation that combines learning capabilities with logic descriptions of the images that are generated. The generation problem is translated into a 
constrained satisfaction problem, where each constraint forces the generated image to have some predefined feature.
A main advantage of this approach is to decouple the logic description level from the generative models. The logic layer is architecture agnostic, allowing to inject into the logic layers any generator model based on deep learning.
In particular, expressing the task using logic knowledge allows to easily extend the involved classes to additional translation categories as well as yielding an easier to understand learning scheme. The translations are then interleaved and jointly learned using the constraints generated by the framework that allow to obtain truly realistic images on different translation types.


Integration of learning and logic reasoning has been studied in the past few years, but no framework emerged as generic interface layer. For example, Minervini et al.~\cite{minervini2017adversarial} corrects the inconsistencies of an adversarial learner but the employed methodology is limited in terms of scope and defined ad-hoc for the task.
A fuzzy generalization of First Order Logic is used both by Hu et al.~\cite{hu2016harnessing} and Logic Tensor Networks~\cite{serafini2016learning} to integrate logic and learning, but both approaches are limited to universally quantified FOL clauses with specific forms.
Another line of research~\cite{rocktaschel2015injecting,demeester2016lifted} attempts at using logical background knowledge to improve the embeddings for Relation Extraction. Also these works are based on ad-hoc solutions that lack a common declarative mechanism that can be easily reused.
Markov Logic Networks (MLN)~\cite{richardson2006markov} and Probabilistic Soft Logic (PSL)~\cite{kimmig2012short,bach2015hinge} are two probabilistic logics, whose parameters are trained to determine the strength of the available knowledge in a given universe. MLN and PSL with their corresponding implementations have received lots of attention but they provide a shallow integration with the underlying learning processes working on the low-level sensorial data. In MLN and PSL, a low-level learner is trained independently, then frozen and stacked with the AI layer providing a higher-level inference mechanism. The framework proposed in this paper instead allows to directly improve the underlying learner, while also providing the higher-level integration with logic.
TensorLog~\cite{cohen2016tensorlog} is a recent framework to reuse the deep-learning infrastructure of TensorFlow (TF) to perform probabilistic logical reasoning. However, TensorLog is limited to reasoning and does not allow to optimize the learners while performing inference.

This paper utilizes a novel framework, called LYRICS~\cite{marra2019lyrics} (Learning Yourself Reasoning and Inference with ConstraintS)\footnote{URL: https://github.com/GiuseppeMarra/lyrics .}, which is a TensorFlow ~\cite{abadi2016tensorflow} environment based on a declarative language for integrating prior knowledge into machine learning. The proposed language generalizes frameworks like Semantic Based Regularization~\cite{diligenti2012bridging,diligenti2015semantic} to any learner trained using gradient descend.
The presented declarative language provides a uniform platform to face both learning and inference tasks by requiring the satisfaction of a set of rules on the domain of discourse. The presented mechanism provides a tight integration of learning and logic as any computational graph can be bound to a FOL predicate.
In the experimental section, an image-to-image task is formulated using logic including adversarial tasks with cycle consistency. The declarative approach allows to easily interleave and jointly learn an arbitrary number of translation tasks.


\section{Constrained Learning and Reasoning}
\label{sec:clare}

\begin{table}[b]
	\centering
	\begin{tabular}{|c|c|c|c|}
		\hline
		\diagbox{formula}{t-norm} & {\bf G$\ddot{\mbox{o}}$del} & {\bf \L ukasiewicz} & {\bf Product} \\
		\hline
		$\neg x$ & $1-x$ & $1-x$ & $1-x$ \\
		\hline
		$x\wedge y$ & $\min\{x,y\}$ & $\max\{0,x+y-1\}$ & $x\cdot y$ \\
		\hline
		$x\vee y$ & $\max\{x,y\}$ & $\min\{1,x+y\}$ & $x+y-x\cdot y$ \\
		\hline
		$x\Rightarrow y$ & $x\leq y?1:y$ & $\min\{1,1-x+y\}$ & $x\leq y?1:y/x$ \\
		\hline
		$x\Leftrightarrow y$ & $x=y?1:\min\{x,y\}$ & $1-|x-y|\}$ & $x=y?1:\min\{x/y,y/x\}$ \\
		\hline
	\end{tabular}
	\vspace{0.1cm}
	\caption{Fundamental t-norms and their algebraic semantics.}
	\label{tab:tnorms}
\end{table}

In this paper, we consider a unified framework where both learning and inference tasks can be seen as constraint satisfaction problems. In particular, the constraints are assumed to be expressed by First-Order Logic (FOL) formulas and implemented in LYRICS, a software we developed converting automatically FOL expressions into TensorFlow computational graphs.

Given a set of task functions to be learned, the logical formalism allows to express high-level statements among the outputs of such functions. For instance, given a certain dataset, if any pattern $x$ has to belong to either a class $A$ or $B$, we may impose that $\forall x:\,f_A(x) \lor f_B(x)$ has to hold true, where $f_A$ and $f_B$ denote two classifiers. As shown in the following of this section, there are several ways to convert FOL into real-valued functions. Exploiting the fuzzy generalization of FOL originally proposed by Novak~\cite{novak1987first}, any FOL knowledge base is translated into a set of real-valued constraints by means of fuzzy logic operators. A \emph{t-norm fuzzy logic} \cite{hajek1998} can be used to transform these statements into algebraic expressions, where a t-norm is a commutative, monotone, associative $[0,1]$-valued operation that models the logical AND. Assuming to convert the logical negation $\neg x$ by means of $1-x$, the algebraic semantics of the other connectives is determined by the choice of a certain t-norm. Different t-norm fuzzy logics have been proposed in the literature and we report in Table~\ref{tab:tnorms} the algebraic operations corresponding to the three fundamental continuous t-norm fuzzy logics, G$\ddot{\mbox{o}}$del, \L ukasiewicz and Product logic. In the following, we will indicate by $\Phi(f(\cal X))$ the algebraic translation of a certain logical formula involving the task functions collected in a vector $f$ and by ${\cal X}$ the available training data.

The constraints are aggregated over a set of data by means of FOL quantifiers. In particular, the universal and existential quantifiers can be seen as a logic AND and OR applied over each grounding of the data, respectively.
Therefore, different quantifiers can be obtained depending on the selection of the underlying t-norm. For example, for a given logic expression $E\big(f({\cal X})\big)$ using the function outputs $f({\cal X})$ as atoms, the product t-norm defines:
\begin{equation}\label{eq:forall}
\forall x_i\, E\big(f({\cal X})\big) \longrightarrow \displaystyle\prod_{x_i \in {\cal X}_i}  \Phi_E\big(f({\cal X})\big) \ ,
\end{equation}
where ${\cal X}_i$ denotes the available sample for the $i$-th task function $f_i$.

In the same way, the expression of the existential quantifier when using the G$\ddot{\mbox{o}}$del t-norm becomes the \textit{maximum} of the expression over the domain of the quantified variable:
\[
\exists x_i\,E\big(f({\cal X})\big) \longrightarrow 
\displaystyle\max_{x_i \in {\cal X}_i} \; \Phi_E\big(f({\cal X}) \big) \ .
\]
Once the translation of the quantifiers are defined, they can be arbitrarily nested and combined in more complicated expressions.


The conversion of formulas into real-valued constraints is carried out automatically in the framework we propose. Indeed, LYRICS takes as input the expressions defined using a declarative language and builds the constraints once we decide the conversion functions to be exploited.
This framework is very general and it accommodates learning from examples as well as the integration with FOL knowledge. In general terms, the learning scheme we propose can be formulated as the minimization of the following cost function:
\def\Bf{\mbox{\boldmath $f$}}
\begin{equation}
\begin{array}{rcl}
C(f( {\cal X} )) &=& \displaystyle\sum_{h=1}^H \lambda_h  \mathcal{L}\Big(\Phi_h \big(f({\cal X})\big) \Big) \ ,
\end{array}
\label{eq:empirical_objective_function}
\end{equation}
where $\lambda_h$ denotes the weight for the $h$-th logical constraint and the function $\mathcal{L}$ represents any monotonically decreasing transformation of the constraints conveniently chosen according to the problem under investigation. In particular, in this paper we exploit the following mappings
\begin{equation}
\begin{array}{l}
\label{eq:L}
{\bf (a)}\;\;\mathcal{L}\Big(\Phi_h \big(f({\cal X})\big) \Big)=1-\Phi_h \big(f({\cal X})\big),\\
{\bf (b)}\;\;\mathcal{L}\Big(\Phi_h \big(f({\cal X})\big) \Big)=-\log\Big(\Phi_h \big(f({\cal X})\big)\Big) \ .
\end{array}
\end{equation}

When the mapping defined in Equation~\ref{eq:L}-{\bf (b)} is applied to an universally quantified formula as in Equation~\ref{eq:forall}, it yields the following constraint:
\[
\mathcal{L}\left(
\displaystyle\prod_{x_i \in {\cal X}_i}  \Phi_E\big(f({\cal X})\big)\right) = -\log
\left(\displaystyle\prod_{x_i \in {\cal X}_i}  \Phi_E\big(f({\cal X})\big)\right) = \displaystyle\sum_{x_i \in {\cal X}_i} -\log\left(
\displaystyle \Phi_E\big(f({\cal X})\big)
\right) \ ,
\]
that corresponds to a generalization to generic fuzzy-logic expressions of the cross-entropy loss, which is commonly used to force the fitting of the supervised data for deep learners.


\begin{example}[From logic formulas to constraints]
Let us consider the rule 
\[
\forall x \forall y ~ Married(x,y)\Rightarrow (Republican(x)\Leftrightarrow Republican(y)) 
\]
where $Republican$ and $Married$ are a unary and a binary predicates indicating if a certain person $x$ votes for a republican and if $x$ is married with a certain person $y$, respectively. The rule states that, if two persons are married, then they vote for the same party. From a learning point of view, enforcing such a rule allows us to exploit the manifold defined by the predicate $Married$ (possibly known) to improve the classification performance about $Republican$ predicate by correlating the predictions of married pairs.
In this case, the input of the predicates can be any vector of features representing a person (e.g. pixel of images, personal data), while the predicates are generally implemented as deep neural models (e.g. a convolutional neural network). 
The rule can be converted into a continuous loss function using e.g. the Product t-norm as reported in table~\ref{tab:tnorms} and the previously reported semantics for the quantifiers:
\[
\prod_{x,y \in {\cal X}} \min\left\{1, \frac{\min\{f_R(x)/f_R(y),f_R(y)/f_R(x)\}}{f_M(x,y)}\right\} \ ,
\]
where $f_R,f_M$ are the functions approximating the predicates $Republican$ and $Married$, respectively and $\cal X$ is the set of patterns representing the available sample of people\footnote{For simplicity we do not consider here the case $f_R(x)=f_R(y)=0$.}. The corresponding loss is obtained by applying Equation~\ref{eq:L}-{\bf (b)}:
\[
\sum_{x,y \in {\cal X}} \max\left\{0,-\log\left( \frac{\min\{f_R(x)/f_R(y),f_R(y)/f_R(x)\}}{f_M(x,y)}\right)\right\} \ .
\]
\end{example}


\section{Generative Learning with Logic}
\label{sec:generative}
This section shows how the discriminative and generative parts of an image-to-image translation system can be formulated by merging logic and learning, yielding a more understandable and easier to extend setup.

Let us assume to be given a set of images $\mathcal{I}$. There are two components of a translator framework. First, a set of \textit{generator} functions $g_{j}: \mathcal{I} \rightarrow \mathcal{I}$, which take as input an image representation and generate a corresponding image in the same output domain, depending on the semantics given to the task. Second, a set of \textit{discriminator} functions $d_i: \mathcal{I} \rightarrow [0,1]$ determining whether an input image $x\in \mathcal{I}$ belongs to class $i$ (i.e. stating if an image has got or not a given property) and, thus, they must be intended in a more general way than in traditional GANs. Interestingly, all learnable FOL functions (i.e. functions mapping input elements into an output element) can be interpreted as generator functions and all learnable FOL predicates (i.e. functions mapping input elements into a truth value) can be interpreted as discriminator functions.

The {\bf discriminator training} corresponds to enforcing the fitting of the supervised examples as:
\begin{equation}\label{eq:discr1}
 \forall x\, S_i(x) \Rightarrow d_i(x), ~~ i = 1,2,\ldots\ 
\end{equation}
 where $S_i(x)$ is a given function returning true if and only if an image is a positive example for the $i$-th discriminator. These constraints allow to transfer the knowledge provided by the supervision (i.e. the $S_i(x)$) inside the discriminators, which play a similar role. However, $d_i(x)$ functions are differentiable and can be exploited to train the generators functions.
To this end, assuming that a given function has to generate an image with a certain property, we can force the corresponding discriminator function for such a property to positively classify it.

The {\bf generator training} for the $j$-th class can be performed by enforcing the generator to produce images that look like images of class $j$, this can be compactly expressed by a rule:
\begin{equation}\label{eq:gen1}
\forall x\, d_j(g_j(x)), ~~ j = 1,2,\ldots
\end{equation}
The logical formalism provides a simple way to describe complex behavior of generator functions by  interleaving multiple positive or negative discriminative atoms (i.e $d_i(g(x))$). 
By requiring that a given image should be classified as realistic, the GAN framework implements a special case of these constraints, where the required property is the similarity with real images.

Cycle consistency~\cite{zhu2017unpaired} is also commonly employed to impose that by translating an image from a domain to another one and then translating it back to the first one, we should recover the input image. Cycle consistency allows to further restrict the number of possible translations. Assuming the semantic of the $i$-th generator is to generate images of class $i$, {\bf cycle consistency} can be naturally formulated as:
\begin{equation}\label{eq:cycle}
\forall x ~ S_i(x) \Rightarrow g_{i}(g_{j}(x)) = x~~ i=1,2,\ldots, ~~ j=1,2,\ldots
\end{equation}
Clearly, in complex problems, the chain of functions intervening in these constraints can be longer.

\begin{figure}[th]
	\centering
	\includegraphics[width=0.3\linewidth]{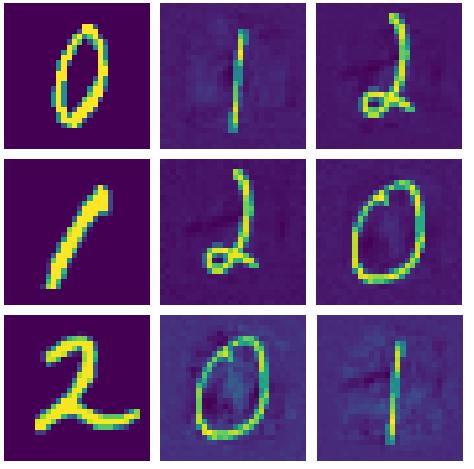}
	\caption{The first column pictures represents the input images. The second and third column pictures show the outputs of the functions \texttt{next} and \texttt{previous}, respectively, computed on the input image.}
	\label{fig:generation}
\end{figure}

The images in different domains are typically required to share the same latent space. Let us indicate $e:\mathcal{I} \rightarrow \mathbb{R}^n$ an encoding function mapping the image into a latent space. This encoding function must be jointly learned during the learning phase. In this special case, the generators must be re-defined as decoder functions taking as input the latent representation of the images, namely:
$g_{j}: \mathbb{R}^n \rightarrow \mathcal{I}$. The {\bf auto-encoding} constraints can be expressed using FOL as follows:
\begin{equation}
\label{eq:identity}
\forall x~ S_i(x)\Rightarrow g_{i}(e(x)) = x, ~~ i=1,2,\ldots
\end{equation}

Up to now, the described constraints are very general and they can be exploited in almost all generative translation tasks. However, the logical formalism (and the LYRICS environment) allows the enforcement of any complex available knowledge about the task at hand. We will see some examples in the following experiment.


\subsubsection{Next and Previous Digits Generation}
As a toy example, we show a task in which we are asked to learn two generative functions, $next$ and $previous$, which, given an image of a $0,1,2$ digit, will produce an image of the next and previous digit, respectively.
In order to give each image a next and a previous digit in the chosen set, a circular mapping was used such that $0$ is the next digit of $2$ and $2$ is the previous digit of $0$.
The functions $next$ and $previous$ are implemented by feedforward neural networks with 50 neurons and 1 hidden layer. Since the output of such functions are still images, the output size of the networks is equal to the input size.
A $1$-hidden layer RBF with a $3$-sized softmax output layer is used to implement the $zero$, $one$ and $two$ discriminators bound to the three outputs of the network, respectively. The RBF model, by constructing closed decision boundaries, allows the generated images to resemble the input ones. 
Finally, let $isZero$, $isOne$ and $isTwo$ be three given functions, defined on the input domain, returning $1$ only if an image is a $0$, $1$ or $2$, respectively. They play the role of the $S_i(x)$ in the general description.

The idea behind this task is to learn generative functions without giving any direct supervision to them, but simply requiring that the generation is consistent with the classification performed by some jointly learned classifiers.
The problem can be described by the following constraints to learn the discriminators
\[
\forall x\,isZero(x) \Rightarrow zero(x), \quad \forall x\,isOne(x)  \Rightarrow one(x),  \quad
\forall x\,isTwo(x)  \Rightarrow two(x)  
\]
and the following constraints to express that the generation functions are constrained to return images which are correctly recognized by the discriminators.
\begin{equation*}
\begin{array}{l}
\forall x~zero(x) \Rightarrow one(next(x)) \land two(previous(x))  \\
\forall x~one(x)  \Rightarrow two(next(x)) \land zero(previous(x)) \\
\forall x~two(x)  \Rightarrow zero(next(x)) \land one(previous(x))
\end{array}
\end{equation*}
In addition, in order to force the generated images to be similar to at least one digit in the domain, we enforce the following constraints:
\begin{equation*}
\begin{array}{l}
\forall x~\exists y~(isZero(x) \land isOne(y)) \Rightarrow next(x) = y \\
\forall x~\exists y~(isZero(x) \land isTwo(y))\Rightarrow previous(x) = y \\ 
\forall x~\exists y~(isOne(x) \land isTwo(y))\Rightarrow next(x) = y \\
\forall x~\exists y~(isOne(x) \land isZero(y))\Rightarrow previous(x) = y \\
\forall x~\exists y~(isTwo(x) \land isZero(y))\Rightarrow next(x) = y \\
\forall x~\exists y~(isTwo(x) \land isOne(y))\Rightarrow previous(x) = y \ .
\end{array}
\end{equation*}
Finally, the cycle consistency constraints can be expressed by:
\[
\forall x\, next(previous(x))  = x \qquad
\forall x\, previous(next(x))  = x \ .
\]

We test this idea by taking a set of around $15000$ images of handwritten characters, obtained extracting only the $0$, $1$ and $2$ digits from the MNIST dataset. The above constraints have been expressed in LYRICS and the model computational graphs have been bound to the predicates.
Figure~\ref{fig:generation} shows an example of image translation using this schema, where the image on the left is an original MNIST image and the two right images are the output of the $next$ and $previous$ generators.


Before proceeding, we want to dwell on the possibilities of this approach after an example has been provided. The declarative nature of the logical formalism and its subsequent translation into real-valued constraints, exploited as loss functions of an optimization problem, enable the construction of very complex generative problems by means of only an high-level semantic description. By exploiting models inherited from the literature, a final user is allowed to face the most different problems with the minimum implementation effort.

In the following section, we show a real image-to-image translation task applying the general setup described in this section, including auto-encoders, GANs and cycle consistency. The declarative nature of the formulation makes very easy to add an arbitrary number of translation problems and it allows to easily learn them jointly.

\section{Experiments on Image Translation}
\label{sec:gan}

\begin{figure*}[t]
	\includegraphics[width=0.98\textwidth]{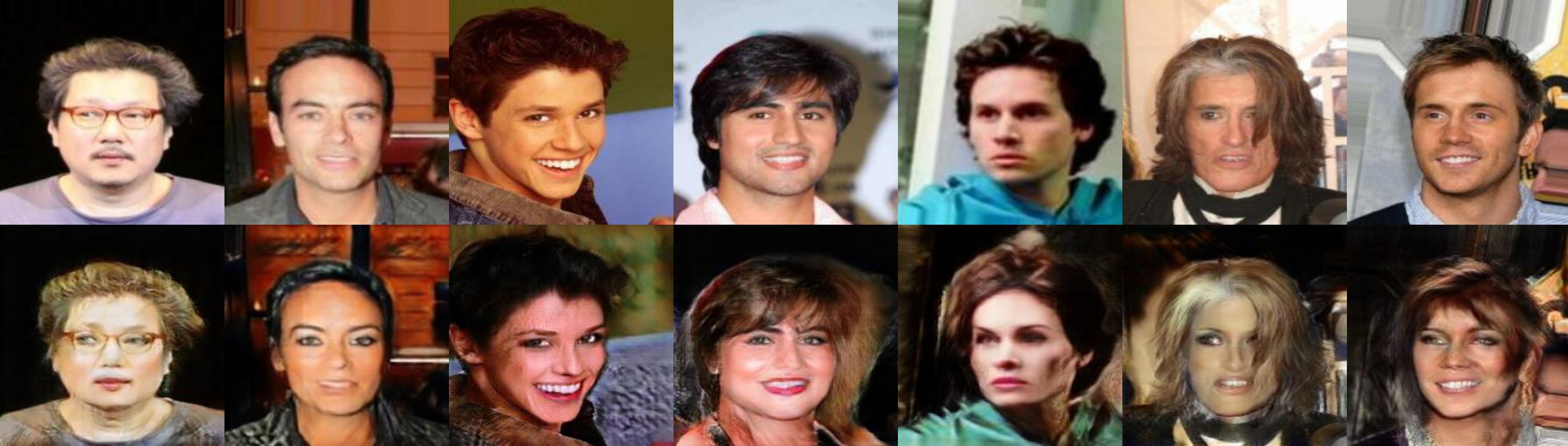}
	\caption{\textbf{Face Gender Translation: male to female.} The top row shows input male images, the bottom row shows the correspondent generated female images.}
	\label{fig:m2f}
\end{figure*}
\begin{figure*}[t]
	\includegraphics[width=0.98\textwidth]{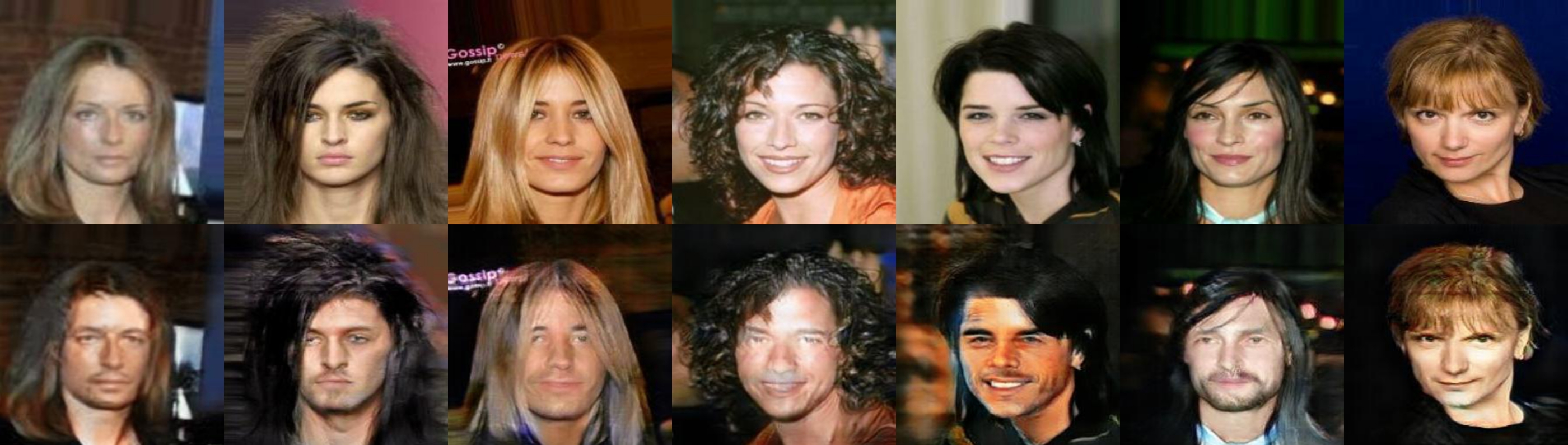}
	\caption{\textbf{Face Gender Translation: female to male.} The top row shows input female images, the bottom row shows the correspondent generated male images.}
	\label{fig:f2m}
\end{figure*}

UNIT translation tasks assume that there are no pairs of examples showing how to translate an image into a corresponding one in another domain.
Combining auto-encoders with GANs is the state-of-the-art solution for tackling UNIT generation problems~\cite{zhu2017unpaired,liu2016coupled,liu2017unsupervised}.
In this section, we show how this adversarial setting can be naturally described and extended by the proposed logical and learning framework. Furthermore, we show how the logical formulation allows a straightforward extension of this application to a greater number of domains.

The CelebFaces Attributes dataset~\cite{liu2015faceattributes} was used to evaluate the proposed approach, where celebrities face images are labeled with various attributes gender, hair color, smiling, eyeglasses, etc.
Images are defined as 3D pixel tensors with values belonging to the $[0,1]$ interval. The first two dimensions represent width and height coordinates while the last dimension indexes among the RGB channels. 

In particular, we used the \emph{Male} attribute to divide the entire dataset into the two input categories, namely male and female images. In the following $S_M(x)$ and $S_F(x)$ (such that $\forall x ~ S_F(x) \Leftrightarrow \lnot S_M(x)$) are two given predicates holding true if and only if an image $x$ is or is not tagged with the \emph{male} tag.
Let $e$ be an encoding function mapping images into the the latent domain ${\mathcal Z}=\mathbb{R}^n$. The encoders are implemented as multilayer convolutional neural networks with resblocks~\cite{he2016deep}, leaky-ReLU activation functions and instance normalization at each layer (see \cite{liu2017unsupervised} for a detailed description of the architecture).
The generative functions $g_M$ and $g_F$ map vectors of the domain $\mathcal Z$ into images. These functions are implemented as multilayer transposed convolutional neural networks (also called “deconvolutions”) with resblocks, leaky-ReLU activation functions and instance normalization at each layer. To implement the shared latent space assumption, $g_M$ and $g_F$ share the parameters of the first layer.

The functions $d_M$ and $d_F$ 
are trained to discriminate whether an image is real or it has been generated by the $g_M$ and $g_F$ generator functions. For example, if $x$ and $y$ are two images such that $S_M(x), S_F(y)$ hold true, then $d_M(x)$ should return $1$ while $d_M(g_M(e(y)))$ should return $0$.

The problem can be described by the logical constraints that have been introduced in a general form in Section \ref{sec:generative} and that the encoding and generation functions need to satisfy. First, Equation~\ref{eq:identity} is used to enforce the encoder and generator of the same domain to be circular, that is to map the input into itself:
\begin{align}
\forall x ~ S_M(x) \Rightarrow g_M(e(x)) = x \label{eq:l11} \\
\forall x  ~ S_F(x) \Rightarrow g_F(e(x)) = x \label{eq:l12}
\end{align}
where the equality operator comparing two images in Equations \ref{eq:l11} and \ref{eq:l12} is bound to a continuous and differentiable function computing a pixel by pixel similarity between the images, defined as $1 -\tanh( \frac{1}{P}\sum_p |x_p - y_p|)$ where $x_p$ and $y_p$ are the $p$-th pixel of the $x$ and $y$ images and $P$ is the total number of pixels.

Cycle consistency is also imposed as described by the Equation~\ref{eq:cycle}.
\begin{align}
\forall x ~ S_M(x) \Rightarrow g_M(e(g_F(e(x))) = x	\label{eq:cycle1} \\
\forall x ~ S_F(x) \Rightarrow g_F(e(g_M(e(x))) = x \label{eq:cycle2}
\end{align}
where the same equality operator is used to compare the images.

Finally, according to the Equation~\ref{eq:gen1}, the generated images must fool the discriminators so that they will be detected as real ones as:
\begin{align}
\forall x ~ S_M(x) \Rightarrow  d_F(g_F(e(x)))\label{eq:adv_g1}\\
\forall x ~ S_F(x)  \Rightarrow  d_M(g_M(e(x)) )\label{eq:adv_g2}
\end{align}

On the other hand, the discriminators must correctly discriminate real images from generated ones by the satisfaction of the following constraints, as stated by Equation~\ref{eq:discr1}:
\begin{align}
\forall x ~ S_M(x) \Rightarrow  d_M(x) \land \lnot d_F(g_F(e(x)))\label{eq:adv_d1}\\
\forall x ~ S_F(x) \Rightarrow  d_F(x) \land \lnot d_M(g_M(e(x))) \label{eq:adv_d2}
\end{align}

Using logical constraints allows us to give a clean and easy formulation of the adversarial setting. These constraints force the generation function to generate samples that are categorized in the desired class by the discriminator. Moreover, the decoupling between the models, used to implement the functions and which can be inherited from the previous literature, and the description of the problem makes really straightforward to extend or transfer this setting.

We implemented this mixed logical and learning task using LYRICS. The Product t-norm was selected to define the underlying fuzzy logic problem. This selection of the t-norm is particularly suited for this task because, as shown earlier, it defines a cross-entropy loss on the output of the discriminators, which is the loss that was used to train these models in their original setup. The $e$, $g_M$, $g_F$ functions are trained to the satisfaction of the constraints defined in  \Cref{eq:l11,eq:l12,eq:cycle1,eq:cycle2,eq:adv_g1,eq:adv_g2}, while $d_M$ and $d_F$ are trained to satisfy  \Cref{eq:adv_d1,eq:adv_d2}. 
Weight learning for the models was performed used the Adam optimizer with a fixed learning rate equal to $0.0001$. Some male-to-female and female-to-male translations are shown in figures \ref{fig:m2f} and \ref{fig:f2m} respectively.

\subsubsection{Adding Eyeglasses}

\begin{figure}[t]
	\includegraphics[width=0.98\textwidth]{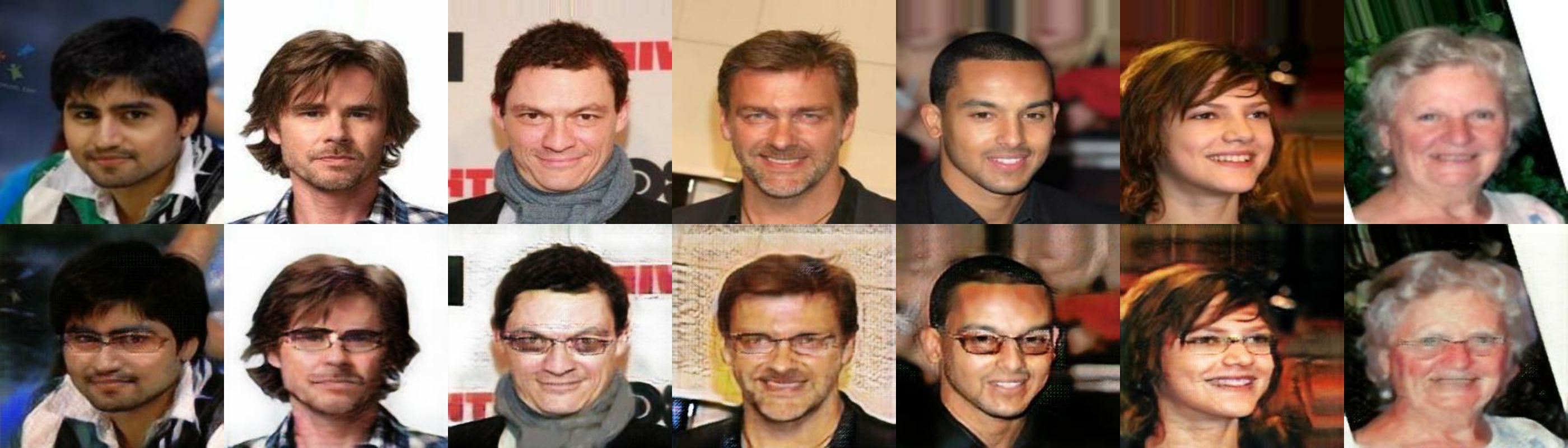}
	\caption[Face Gender Translation: male/female to eyeglasses]{\textbf{Face Gender Translation: male/female to eyeglasses.} The top row shows input male/female images whereas the bottom row shows the correspondent generated  faces with eyeglasses.}
	\label{fig:eyeglasses}
\end{figure}

Given this setting, we can integrate a third domain in the overall problem adding the corresponding constraints for this class. Let $S_E(x)$ be a given predicate holding true if and only if an image $x$ is tagged with the \emph{eyeglasses} tag in the dataset. Let $g_E(x)$  be the corresponding generator and $d_E(x)$ the corresponding discriminator for this property. The same network architectures of the previous description are employed to implement $d_E$ and $g_E$. The addition of this third class requires to add the following constraints for the generators, to be integrated with the male and female classes,
\begin{align*}
\forall x& ~ S_M(x) \Rightarrow d_E(g_E(e(x)))\\
\forall x& ~ S_F(x) \Rightarrow d_E(g_E(e(x)))\\
\forall x& ~ S_E(x) \Rightarrow g_E(e(x)) = x  \\
\forall x& ~ S_M(x) \wedge S_E(x) \Rightarrow d_E(g_F(e(x))) \\ 
\forall x& ~ S_F(x) \wedge S_E(x) \Rightarrow d_E(g_M(e(x))) \\ 
\forall x& ~ S_M(x) \wedge S_E(x) \Rightarrow g_E(e(g_F(e(x))) = g_F(e(x)) \\
\forall x& ~ S_F(x) \wedge S_E(x) \Rightarrow g_E(e(g_M(e(x))) = g_M(e(x)) \\
\forall x& ~ S_M(x) \wedge \neg S_E(x) \Rightarrow g_M(e(g_E(e(x))) = g_E(e(x)) \\
\forall x& ~ S_F(x) \wedge \neg S_E(x) \Rightarrow g_F(e(g_E(e(x))) = g_E(e(x)) 
\end{align*}
and  to add the following for the discriminator:
\begin{align*}
\forall x& ~ S_E(x) \Rightarrow  d_E(x)\\
\forall x& ~ S_M(x)\wedge\neg S_E(x) \Rightarrow \neg d_E(g_E(e(x)))\\
\forall x& ~ S_F(x)\wedge\neg S_E(x) \Rightarrow \neg d_E(g_E(e(x)))
\end{align*}
We note that in this case, the class eyeglasses is not mutually exclusive neither with male nor female class. This is the reason why we have to consider some constraints with a conjunction on premises. In addition, we have to distinguish how the male and female generators behave in presence of the attribute eyeglasses. In particular we enforce that translating a gender attribute does not affect the presence of eyeglasses. Figure~\ref{fig:eyeglasses} shows some examples of the original face images, and the corresponding generated images of the faces with added eyeglasses.

As we already said, the proposed approach is very general and can be exploited to manage possibly several attributes in a visual generation task combining a high-level logical description with deep neural networks.
The most distinguishing property is the flexibility of describing new generation problems by simple
logic descriptions, which leads to attack very different problems. Instead of looking for specific hand-crafted cost functions, the proposed approach offers a general scheme for their construction that arises from the
t-norm theory. Moreover, the interleaving of
different image translations tasks allows us to
accumulate a knowledge base that can
dramatically facilitate the construction of new
translation tasks. The experimental results
shows the flexibility of the proposed approach,
which makes it possible to deal with realistic
face translation tasks.

\section{Conclusions}

\label{sec:conclusion}

This paper shows a new general approach to visual generation combining logic descriptions of the target to be generated with deep neural networks.
The most distinguishing property is the flexibility of describing new generation problems by simple logic descriptions, which leads to attack very different problems. Instead of looking for specific hand-crafted cost functions, the proposed approach offers a general scheme for their construction that arises from the t-norm theory. Moreover, the interleaving of different image translations tasks allows to
accumulate a knowledge base that can dramatically facilitate the construction of new
translation tasks. The experimental results shows the flexibility of the proposed approach,
which makes it possible to deal with realistic face translation tasks.

%

 \bibliographystyle{splncs04}
 \bibliography{references}

\begin{thebibliography}{10}
\providecommand{\url}[1]{\texttt{#1}}
\providecommand{\urlprefix}{URL }
\providecommand{\doi}[1]{https://doi.org/#1}

\bibitem{abadi2016tensorflow}
Abadi, M., Barham, P., Chen, J., Chen, Z., Davis, A., Dean, J., Devin, M.,
  Ghemawat, S., Irving, G., Isard, M., et~al.: Tensorflow: A system for
  large-scale machine learning. In: OSDI. vol.~16, pp. 265--283 (2016)

\bibitem{bach2015hinge}
Bach, S.H., Broecheler, M., Huang, B., Getoor, L.: Hinge-loss markov random
  fields and probabilistic soft logic. arXiv preprint arXiv:1505.04406  (2015)

\bibitem{cohen2016tensorlog}
Cohen, W.W.: Tensorlog: A differentiable deductive database. arXiv preprint
  arXiv:1605.06523  (2016)

\bibitem{demeester2016lifted}
Demeester, T., Rockt{\"a}schel, T., Riedel, S.: Lifted rule injection for
  relation embeddings. arXiv preprint arXiv:1606.08359  (2016)

\bibitem{diligenti2012bridging}
Diligenti, M., Gori, M., Maggini, M., Rigutini, L.: Bridging logic and kernel
  machines. Machine learning  \textbf{86}(1),  57--88 (2012)

\bibitem{diligenti2015semantic}
Diligenti, M., Gori, M., Sacc{\`a}, C.: Semantic-based regularization for
  learning and inference. Artificial Intelligence  (2015)

\bibitem{goodfellow2014generative}
Goodfellow, I., Pouget-Abadie, J., Mirza, M., Xu, B., Warde-Farley, D., Ozair,
  S., Courville, A., Bengio, Y.: Generative adversarial nets. In: Advances in
  neural information processing systems. pp. 2672--2680 (2014)

\bibitem{hajek1998}
Hajek, P.: The Metamathematics of Fuzzy Logic. Kluwer (1998)

\bibitem{he2016deep}
He, K., Zhang, X., Ren, S., Sun, J.: Deep residual learning for image
  recognition. In: Proceedings of the IEEE conference on computer vision and
  pattern recognition. pp. 770--778 (2016)

\bibitem{hu2016harnessing}
Hu, Z., Ma, X., Liu, Z., Hovy, E., Xing, E.: Harnessing deep neural networks
  with logic rules. arXiv preprint arXiv:1603.06318  (2016)

\bibitem{kimmig2012short}
Kimmig, A., Bach, S., Broecheler, M., Huang, B., Getoor, L.: A short
  introduction to probabilistic soft logic. In: Proceedings of the NIPS
  Workshop on Probabilistic Programming: Foundations and Applications. pp.~1--4
  (2012)

\bibitem{li2017alice}
Li, C., Liu, H., Chen, C., Pu, Y., Chen, L., Henao, R., Carin, L.: Alice:
  Towards understanding adversarial learning for joint distribution matching.
  In: Advances in Neural Information Processing Systems. pp. 5501--5509 (2017)

\bibitem{liu2017unsupervised}
Liu, M.Y., Breuel, T., Kautz, J.: Unsupervised image-to-image translation
  networks. In: Advances in Neural Information Processing Systems. pp. 700--708
  (2017)

\bibitem{liu2016coupled}
Liu, M.Y., Tuzel, O.: Coupled generative adversarial networks. In: Proceedings
  of the 30th International Conference on Neural Information Processing
  Systems. pp. 469--477. Curran Associates Inc. (2016)

\bibitem{liu2015faceattributes}
Liu, Z., Luo, P., Wang, X., Tang, X.: Deep learning face attributes in the
  wild. In: Proceedings of International Conference on Computer Vision (ICCV)
  (December 2015)

\bibitem{marra2019lyrics}
Marra, G., Giannini, F., Diligenti, M., Gori, M.: Lyrics: a general interface
  layer to integrate ai and deep learning. arXiv preprint arXiv:1903.07534
  (2019)

\bibitem{minervini2017adversarial}
Minervini, P., Demeester, T., Rockt{\"a}schel, T., Riedel, S.: Adversarial sets
  for regularising neural link predictors. arXiv preprint arXiv:1707.07596
  (2017)

\bibitem{novak1987first}
Nov{\'a}k, V.: First-order fuzzy logic. Studia Logica  \textbf{46}(1),  87--109
  (1987)

\bibitem{richardson2006markov}
Richardson, M., Domingos, P.: Markov logic networks. Machine learning
  \textbf{62}(1),  107--136 (2006)

\bibitem{rocktaschel2015injecting}
Rockt{\"a}schel, T., Singh, S., Riedel, S.: Injecting logical background
  knowledge into embeddings for relation extraction. In: Proceedings of the
  2015 Conference of the North American Chapter of the Association for
  Computational Linguistics: Human Language Technologies. pp. 1119--1129 (2015)

\bibitem{rosca2017variational}
Rosca, M., Lakshminarayanan, B., Warde-Farley, D., Mohamed, S.: Variational
  approaches for auto-encoding generative adversarial networks. arXiv preprint
  arXiv:1706.04987  (2017)

\bibitem{serafini2016learning}
Serafini, L., Garcez, A.S.d.: Learning and reasoning with logic tensor
  networks. In: AI* IA. pp. 334--348 (2016)

\bibitem{zhu2017unpaired}
Zhu, J.Y., Park, T., Isola, P., Efros, A.A.: Unpaired image-to-image
  translation using cycle-consistent adversarial networks. In: Proceedings of
  the IEEE Conference on Computer Vision and Pattern Recognition. pp.
  2223--2232 (2017)

\end{thebibliography}

\end{document}